\documentclass{article}

\usepackage{arxiv}
\usepackage{multirow}
\usepackage{booktabs}
\usepackage[utf8]{inputenc} 
\usepackage[T1]{fontenc}    
\usepackage{hyperref}       
\usepackage{url}            
\usepackage{booktabs}       
\usepackage{amsfonts}       
\usepackage{nicefrac}       
\usepackage{microtype}      
\usepackage{lipsum}
\usepackage{graphicx}
\graphicspath{ {./images/} }

\title{BNLI: A Linguistically-Refined Bengali Dataset for Natural Language Inference}

\author{
  Farah Binta Haque \\
  Computer Science and Engineering\\
  Brac University\\
  Dhaka, Bangladesh \\
  \texttt{farahhaq2023@gmail.com} \\
   \And
  Md Yasin\thanks{These authors contributed equally to this work.} \\
  Computer Science and Engineering\\
  BRAC University\\
  Dhaka, Bangladesh \\
  \texttt{yasinislam046@gmail.com} \\
  \And
  Shishir Saha\footnotemark[1] \\
  Computer Science and Engineering\\
  BRAC University\\\
  Dhaka, Bangladesh \\
  \texttt{shishirsaha830@gmail.com} \\
  \And
  Md Shoaib Akhter Rafi \\
  Electrical and Electronic Engineering\\
  Bangladesh University of Engineering and Technology\\
  Dhaka, Bangladesh \\
  \texttt{1806114@eee.buet.ac.bd} \\
  \And
  Farig Sadeque \\
  Computer Science and Engineering\\
  BRAC University\\
  Dhaka, Bangladesh \\
  \texttt{farig.sadeque@bracu.ac.bd} \\
}
\makeatletter
\renewcommand{\@date}{}
\makeatother

\begin{document}
\maketitle
\begin{abstract}
Despite the growing progress in Natural Language Inference (NLI) research, resources for the Bengali language remain extremely limited. Existing Bengali NLI datasets exhibit several inconsistencies, including annotation errors, ambiguous sentence pairs, and inadequate linguistic diversity, which hinder effective model training and evaluation. To address these limitations, we introduce BNLI, a refined and linguistically curated Bengali NLI dataset designed to support robust language understanding and inference modeling. The dataset was constructed through a rigorous annotation pipeline emphasizing semantic clarity and balance across entailment, contradiction, and neutrality classes. We benchmarked BNLI using a suite of state-of-the-art transformer-based architectures, including multilingual and Bengali-specific models, to assess their ability to capture complex semantic relations in Bengali text. The experimental findings highlight the improved reliability and interpretability achieved with BNLI, establishing it as a strong foundation for advancing research in Bengali and other low-resource language inference tasks.
\end{abstract}


\section{Introduction}
Natural Language Inference (NLI), also referred to as Recognizing Textual Entailment (RTE), represents a core task in natural language understanding that requires determining whether a given hypothesis can be inferred from a corresponding premise. Over the past decade, large-scale English NLI datasets such as SNLI \cite{bowman2015large}, MultiNLI \cite{williams2018broad}, and ANLI \cite{nie2017shortcut} have become foundational benchmarks for advancing deep language models including BERT, RoBERTa, and LLaMA. These resources have substantially contributed to progress in semantic reasoning and contextual understanding across diverse NLP applications \cite{bowman2015large, williams2018broad, gururangan2018annotation}. However, comparable resources for low-resource languages—particularly Bengali—remain strikingly underdeveloped.

Although Bengali ranks among the top ten most spoken languages globally, Bengali Natural Language Processing (BNLP) still suffers from a scarcity of high-quality annotated corpora for complex tasks such as entailment recognition \cite{aggarwal2022indicxnli, bhattacharjee2022banglabert, kabir2024benllm}. Only a handful of preliminary efforts have been made to construct Bengali NLI datasets; however, most existing corpora suffer from severe shortcomings, including inaccurate translations from English datasets \cite{aggarwal2022indicxnli, bhattacharjee2022banglabert}, inconsistent labeling, syntactic errors, and a lack of semantic and contextual diversity. Consequently, models trained on these resources often exhibit limited generalization, biased performance, and poor interpretability when exposed to real-world Bengali text \cite{kabir2024benllm, faria2024unraveling, mahfuz2025too}. This persistent data deficiency has impeded meaningful advancement in Bengali inference modeling and constrained cross-lingual transfer research.

To bridge this critical gap, we present BNLI—a refined and linguistically curated Bengali Natural Language Inference dataset—specifically designed to facilitate robust evaluation and training of NLI models in Bengali. BNLI was developed through a rigorous multi-stage pipeline that combines manual data curation, human validation, and linguistic refinement to ensure both semantic consistency and syntactic authenticity. Unlike earlier datasets, BNLI emphasizes balanced representation across entailment, contradiction, and neutrality classes and incorporates examples reflecting diverse linguistic structures and contextual nuances present in natural Bengali text.

To establish baselines, we evaluated BNLI using several transformer-based architectures, including multilingual and Bengali-specific models. The experimental results demonstrate that transformer models outperform recurrent approaches by effectively capturing contextual semantics and long-range dependencies in Bengali. These findings underscore the reliability of BNLI as a benchmark and highlight its potential to accelerate progress in Bengali and other low-resource language inference research \cite{rashad2024banglaquad}.

In summary, the main contributions of this study are:

\begin{itemize}
    \item We develop BNLI, a high-quality and linguistically refined Bengali NLI dataset addressing critical flaws in previous resources.
    \item We provide comprehensive benchmark evaluations using multiple transformer-based architectures to establish reliable baselines for future research.
\end{itemize}

By making BNLI publicly available, we aim to promote further exploration in Bengali semantic understanding, encourage resource development for other low-resource languages, and foster inclusive multilingual NLI research.

\section{Literature Review}

\begin{table}[!ht]
 \caption{List of existing NLI Datasets on diverse languages including Bangla.}
  \centering
  \begin{tabular}{lccc}
    \toprule
    \textbf{Dataset} & \textbf{Language} & \textbf{Year}   & \textbf{Sample Size}  \\
    \midrule
      SNLI \cite{bowman2015large} &  English & 2015  & $\sim$570k\\
      MultiNLI \cite{williams2018broad}   & English  & 2017   & $\sim$433k \\
      e-SNLI \cite{camburu2018snli}    &  English &  2018 & $\sim$570k\\
      ANLI \cite{nie2020adversarial}    &    English  & 2018 & $\sim$162k   \\
      SICK \cite{marelli2014semeval}           & English         &  2019-2020     &   $\sim$10k \\
      SciTail \cite{khot2018scitail} &   English       &  2014      & $\sim$27k   \\
      RTE \cite{castillo2010recognizing}   &   English       & 2004-2013       & Not Found    \\
      SCINLI \cite{sadat2022scinli}    &    English      & 2022       & $\sim$107k    \\
      FEVER \cite{thorne2018fever}        &   English    &  2018      & $\sim$185K   \\
      QNLI \cite{swayamdipta2020dataset}       &     English     &   2018     &   $\sim$100K \\
      ARABIC-XNLI \cite{abdelali2024larabench}         &Arabic     &2018     &   $\sim$392k \\
      ARBERT / MARBERT benchmarks \cite{abdul2021arbert} & Arabic &2021 & $\sim$20k\\
     Arabic Natural Language Inference Corpus \cite{obeidat2025arentail} & Arabic & 2019 & $\sim$9k\\
     SANA (Sentiment and NLI Arabic) \cite{abdul2014sana} & Arabic & 2020 & $\sim$7k\\
     OSIAN (Open Source International Arabic News) \cite{zeroual2019osian} &Arabic &2019 & $\sim$1.1M\\
     ViNLI \cite{van2022vinli} & Vietnamese & 2021 & $\sim$7.5k\\
     ViHealthNLI \cite{nguyen2024vihealthnli} & Vietnamese & 2024 & $\sim$5k\\
     VietNLI \cite{bui2025vietx} & Vietnamese & 2023 & $\sim$25k\\
     XNLI (Cross-lingual NLI) \cite{conneau2018xnli} & Multilingual & 2018 & $\sim$112k\\
     INDICXNLI \cite{aggarwal2022indicxnli} & Multilingual & 2022 & $\sim$392k \\
     Multilingual-NLI-26lang-2mil7 \cite{he2020deberta} & Multilingual &2022 & $\sim$2.73M\\
     Cross-Lingual NLI for Low-Resource Languages (MT-NLI) \cite{zhao2022cross} & Multilingual &2023 & $\sim$3.5M\\
     
     SICK-NL \cite{wijnholds2021sicknl} & Dutch &2021 & $\sim$10k\\
     NLI-TR \cite{budur2020data} & Turkish &2020 & $\sim$393k \\
     farsTail \cite{amirkhani2023farstail} & Persian &2020 & $\sim$10k\\
     OCNLI (Original Chinese NLI) \cite{hu2020ocnli} & Chinese & 2020 & $\sim$56k\\
     Polish NLI \cite{ziembicki2024polish} & Polish & 2022 & $\sim$10k\\
     Bangla NLI (BanglaBERT benchmarks) \cite{bhattacharjee2022banglabert} &Bangla &2022 & $\sim$381k\\
     
    \bottomrule
  \end{tabular}
  \label{tab:table}
\end{table}

Table \ref{tab:table} presents an overview of major Natural Language Inference (NLI) datasets developed across various languages over the past two decades. As illustrated, extensive and well-curated resources exist for high-resource languages such as English, Arabic, and Chinese, as well as for several emerging multilingual benchmarks like XNLI, INDICXNLI, and MT-based multilingual corpora. These datasets—ranging from early collections such as RTE and SNLI to large-scale multilingual resources—have substantially driven progress in semantic understanding and cross-lingual inference modeling.

In contrast, Bengali remains severely underrepresented in this landscape. Only a limited number of Bengali NLI datasets have been introduced, and most of them are either small in scale or derived from machine-translated versions of English benchmarks. Moreover, many samples in these datasets suffer from semantic inconsistencies, mistranslations, and contextually incorrect entailment labeling, thereby limiting their reliability for model training and evaluation. This evident disparity underscores the urgent need for a linguistically consistent and semantically validated Bengali NLI resource. To address this gap, we introduce BNLI, a refined and meticulously curated dataset aimed at ensuring semantic integrity, syntactic correctness, and balanced class representation for robust Bengali inference modeling.

\section{Methodology}

\begin{figure*}[!ht]
    \centering
    \includegraphics[width=1\textwidth, trim={12.2cm 3.0cm 12.2cm 3.0cm}, clip]{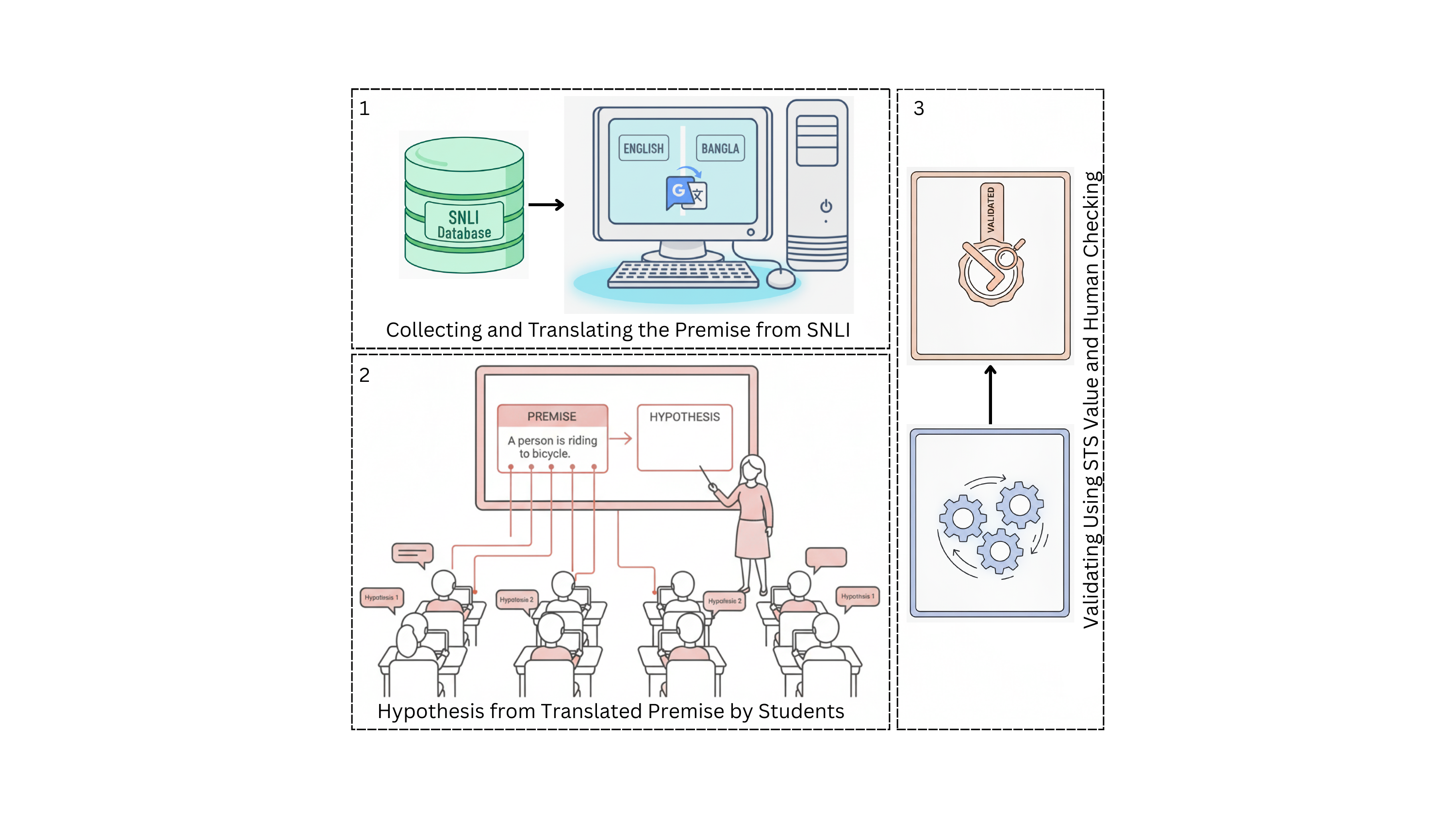}
    \caption{Overall workflow of the proposed multi-stage pipeline adopted for constructing the BNLI dataset, illustrating data collection, linguistic refinement, human validation, and final dataset compilation processes.}
    \label{fig:bnli_method}
\end{figure*}

The BNLI dataset was constructed through a three-stage data collection and refinement process to ensure linguistic accuracy and semantic reliability. The data collection process is illustrated in Fig.~\ref{fig:bnli_method}.

In the first stage, 8885 sentences were selected as premises from the SNLI corpus \cite{bowman2015large}. These sentences were translated into Bengali with careful attention to preserving grammatical integrity and natural sentence flow. In the second stage, we focused on constructing semantically aligned hypotheses for the collected Bengali premises. To generate corresponding hypotheses, 60 native Bengali-speaking students from a reputed university participated in a structured data collection process. Each participant was provided with premise–scene descriptions and tasked with composing hypothesis sentences through Google Forms. This process resulted in 26655 sentence pairs, evenly distributed across the three NLI labels: entailment, contradiction, and neutral.

In the third stage, all collected sentence pairs underwent a comprehensive validation process combining Semantic Textual Similarity (STS) analysis and human evaluation. Each premise–hypothesis pair was translated into English using the Google Translator Python package, and STS scores were computed to assess semantic alignment with the assigned NLI labels. Pairs with inconsistent or outlier scores were flagged for manual inspection. Subsequently, five native Bengali linguists reviewed these pairs to verify grammatical correctness, semantic coherence, and label validity, resolving discrepancies through consensus. The final dataset contains total 23067 refined sentence pairs (Entailment: 7682 pairs, Contradiction: 7696 pairs, Neutral: 7661 pairs). This dual validation approach ensured that the BNLI dataset maintained high linguistic fidelity, semantic consistency, and balanced representation across inference categories.

\section{Result Analysis}

\begin{table*}[!ht]
\centering
\caption{Model-wise performance comparison on BNLI dataset.}
\begin{tabular}{l| c | c|c|c | c|c|c|c}
\hline
\multirow{2}{*}{\textbf{Model}} & 
\multirow{2}{*}{\textbf{\#Params(M)}} &
\multicolumn{3}{c|}{\textbf{Classwise Performance}} &
\multicolumn{4}{c}{\textbf{Average Performance}} \\
\cline{3-5} \cline{6-9}
& & \textbf{Contradiction} & \textbf{Neutral} & \textbf{Entailment} 
& \textbf{Accuracy} & \textbf{Precision} & \textbf{Recall} & \textbf{F1-score} \\
\hline
LSTM & $\sim$10 & 45.12\% & 46.38\% & 41.57\% &44.19\% & 45.06\% & 44.32\% & 44.68\% \\
BERT Base & 110 & 76.42\% & 73.19\% & 64.87\% & 70.90\% &72.45\% & 71.66\% & 71.92\% \\
BERT Large & 340 & 83.18\% & 68.73\% & 53.24\% &67.5\% &68.12\% & 68.44\% & 68.27\% \\
BanglaBERT & 168 & 32.46\% & 11.58\% & 88.12\% &45.4\% &49.37\% & 48.73\% & 44.26\% \\
RoBERTa & 125 & 79.68\% & 74.91\% & 67.43\% &69.66\% &74.65\% & 73.92\% & 74.21\% \\
MultiBERT & 185 & 86.23\% & 75.62\% & 48.91\% &68.75 & 69.58\% & 70.34\% & 69.93\% \\
LLaMA-2 & 7,000 & \textbf{84.76\%} & \textbf{80.35\%} & \textbf{72.41\%} & \textbf{74.56\%} & \textbf{79.51\%} & \textbf{78.92\%} & \textbf{79.13\%} \\
\hline
\end{tabular}
\label{tab:model_performance}
\end{table*}

The results on the BNLI dataset (Table.~\ref{tab:model_performance}) show that transformer-based models substantially outperform traditional architectures. The LSTM baseline performs poorly, reflecting its limited ability to capture semantic nuances. BERT Base and RoBERTa deliver strong and balanced results, with RoBERTa slightly ahead due to improved contextual representations. BanglaBERT shows inconsistent performance across classes, suggesting limitations in domain coverage. MultiBERT performs reasonably well, benefiting from multilingual training. The LLaMA-2 model achieves the highest overall F1-score (79.13\%), demonstrating superior cross-lingual reasoning and robust generalization compared to all other models.

\section{Conclusion}
In this work, we presented BNLI, a high-quality Bengali NLI dataset addressing the limitations of existing resources, including annotation inconsistencies and linguistic imbalance. Through a careful curation and annotation process, BNLI ensures clear semantic distinctions across entailment, contradiction, and neutral classes, making it suitable for reliable model evaluation and training. Our extensive benchmarking with both multilingual and Bengali-specific transformer models demonstrates the dataset’s effectiveness in revealing model strengths and weaknesses, as well as the substantial performance gains achievable with large-scale language models such as LLaMA-2. Overall, BNLI provides a robust foundation for advancing Bengali natural language understanding and inference research, and we anticipate that it will facilitate the development of more accurate, interpretable, and generalizable NLI models for low-resource languages.

\bibliographystyle{jbrv.bst}
\bibliography{references.bib}

\end{document}